# A Knowledge Mining Model for Ranking Institutions using Rough Computing with Ordering Rules and Formal Concept Analysis

D. P. Acharjya[1], and L. Ezhilarasi[2]

[1, 2] School of Computing Sciences and Engineering, VIT University
Vellore, Tamil Nadu, India

**Abstract**
Emergences of computers and information technological revolution made tremendous changes in the real world and provides a different dimension for the intelligent data analysis. Well formed fact, the information at right time and at right place deploy a better knowledge. However, the challenge arises when larger volume of inconsistent data is given for decision making and knowledge extraction. To handle such imprecise data certain mathematical tools of greater importance has developed by researches in recent past namely fuzzy set, intuitionistic fuzzy set, rough Set, formal concept analysis and ordering rules. It is also observed that many information system contains numerical attribute values and therefore they are almost similar instead of exact similar. To handle such type of information system, in this paper we use two processes such as pre process and post process. In pre process we use rough set on intuitionistic fuzzy approximation space with ordering rules for finding the knowledge whereas in post process we use formal concept analysis to explore better knowledge and vital factors affecting decisions.

## 1. Introduction

In real world, invent of computers have created a new space for knowledge mining. Thus, data handling and data processing is of prime importance in the information system. In the hierarchy of data processing, data is the root which transforms into information and further refined to avail it in the form of knowledge. Knowledge mining specifies the collection of the right information at the right level and utilize for the suitable purposes. It is important to understand that, at human level it involves a thought process. These thoughts are responsible to collect the raw data and convert it in to usable data and further to information for an assigned task. The real challenge arises when larger volume of inconsistent data is presented for extraction of knowledge and decision making. The major issue lies in converting large volume of data into knowledge and to use that knowledge to make a proper decision. Current technologies help in obtaining decisions by creating large databases. However, it is failure in many instances because of irrelevant information in the databases. It leads to attribute reduction. Thus, attribute reduction is an important factor for handling such huge data sets by eliminating the superficial data to provide an effective decision. Many methods where proposed to mine rules from the growing data. Most of the tools to mine knowledge are crisp, traditional, deterministic and precise in notion. Real situations are very often the reverse of it. The detailed description of the real system needs detailed data which is beyond the recognition of the human interpretation. This invoked an extension of the concept of crisp sets so as to model imprecise data that enable their modeling intellects. Typically, each model and method presents a particular and single view of data or discovers a specific type of knowledge embedded in the data. To carry on with inconsistent data certain mathematical tools of greater importance got developed in recent past namely fuzzy set [7], intuitionistic fuzzy set [6], rough set [11], formal concept analysis [8], and ordering rules [10] to name a few. Further rough set is generalized to rough sets on fuzzy approximation space [9], rough sets on intuitionistic fuzzy approximation spaces [1]. The notion of rough sets on intuitionistic fuzzy approximation space depends on the concept of intuitionistic fuzzy proximity relation.

In this paper, we propose an integrated knowledge mining model that combines rough set on intuitionistic fuzzy approximation space with ordering of objects and formal concept analysis as a new technique for extracting better knowledge from the available data set. The motivation behind this study is that the two theories aim at different goals and summarize different types of knowledge. Rough computing is used for prediction whereas formal concept analysis is used for description. Therefore, the combination of both leads to better knowledge mining model. However, for completeness of the paper we explain the basic concepts of rough sets in section 2, rough sets on intuitionistic fuzzy approximation spaces in section 3. In section 4, we discuss order information table followed by formal concept analysis in section 5. We present our proposed model in section 6 and is further analyzed in section 7. The paper is concluded in section 8.

## 2. Foundations of Rough Sets

The theory of rough sets has been under continuous development for over few decades and a fast growing group of researchers and practitioners are interested in this methodology. The methodology is concerned with the classificatory analysis of imprecise, uncertain or incomplete data. The main advantage of rough set theory proposed by Pawlak [15] is that it does not need any preliminary or additional information about data. On the other hand, handling vagueness is one of the motivations for proposing the rough set theory [15]. Objects that belong to the same category of the information are indiscernible [12, 13, 14]. Thus information associated with objects of the universe generates an indiscernibility relation and is the basic concept of rough set theory. This fact leads to the definition of a set in terms of lower and upper approximations, where approximation space is the basic notion of the rough set theory.

Let $U$ be a finite nonempty set called the universe. Suppose $R \subseteq U \times U$ is an equivalence relation defined on $U$. Thus, the equivalence relation $R$ partitions the set $U$ into disjoint subsets [15]. Elements of same equivalence class are said to be indistinguishable. Equivalence classes induced by $R$ are called elementary sets or elementary concepts. The pair $(U, R)$ is called an approximation space. Given a target set $X \subseteq U$, we can characterize $X$ by a pair of lower and upper approximations. We associate two subsets $\underline{R}X$ and $\overline{R}X$ called the $R$–Lower and $R$–Upper approximations of $X$ respectively and are given by

$$\underline{R}X = \cup\{Y \in U / R : Y \subseteq X\} \quad (1)$$

and $\overline{R}X = \cup\{Y \in U / R : Y \cap X \neq \phi\} \quad (2)$

The $R$–boundary of $X$, $BN_R(X)$ is given by $BN_R(X) = \underline{R}X - \overline{R}X$. We say $X$ is rough with respect to $R$ if and only if $\underline{R}X \neq \overline{R}X$ or $BN_R(X) \neq \phi$. $X$ is said to be $R$–definable if and only if $\underline{R}X = \overline{R}X$ or $BN_R(X) = \phi$.

## 3. Rough Sets on Intuitionistic Fuzzy Approximation Spaces

As mentioned in the previous section, the basic rough set depends upon equivalence relations. However, such types of relations are rare in information system containing numerical values. This leads to the dropping of the transitivity property of an equivalence relation and making reflexivity and symmetry property lighter. It leads to fuzzy proximity relation and thus rough sets on fuzzy approximation spaces [9]. The fuzzy proximity relation is further generalized to intuitionistic fuzzy proximity relation [6]. Thus rough sets on intuitionistic fuzzy approximation spaces introduced by Tripathy [1] provides a better model over rough set and rough set on fuzzy approximation space. The different properties of rough sets on intuitionistic fuzzy approximation spaces were studied by Acharjya and Tripathy [5]. However, for completeness of the paper we provide the basic notions of rough sets on intuitionistic fuzzy approximation spaces. We use standard notation $\mu$ for membership and $\nu$ for non-membership functions associated with an intuitionistic fuzzy set.

**Definition 3.1** An intuitionistic fuzzy relation $R$ on a universal set $U$ is an intuitionistic fuzzy set defined on $U \times U$.

**Definition 3.2** An intuitionistic fuzzy relation $R$ on U is said to be an intuitionistic fuzzy proximity relation if the following properties hold.

$$\mu_R(x, x) = 1 \text{ and } \nu_R(x, x) = 0 \quad \forall \ x \in U \quad (3)$$

$$\mu_R(x, y) = \mu_R(y, x), \nu_R(x, y) = \nu_R(y, x) \quad \forall \ x, y \in U \quad (4)$$

**Definition 3.3** Let $R$ be an intuitionistic fuzzy (IF) proximity relation on $U$. then for any $(\alpha, \beta) \in J$, where $J = \{(\alpha, \beta) \mid \alpha, \beta \in [0, 1] \text{ and } 0 \leq \alpha + \beta \leq 1\}$, the $(\alpha, \beta)$-cut $'R_{\alpha, \beta}'$ of $R$ is given by

$$R_{\alpha, \beta} = \{(x, y) \mid \mu_R(x, y) \geq \alpha \text{ and } \nu_R(x, y) \leq \beta\}$$

**Definition 3.4** Let $R$ be an IF-proximity relation on $U$. We say that two elements $x$ and $y$ are $(\alpha, \beta)$-similar with respect to $R$ if $(x, y) \in R_{\alpha, \beta}$ and we write $x R_{\alpha, \beta} y$.

**Definition 3.5** Let $R$ is an IF-proximity relation on $U$. We say that two elements $x$ and $y$ are $(\alpha, \beta)$-identical with respect to $R$ for $(\alpha, \beta) \in J$, written as $xR(\alpha, \beta)y$ if and only if $x R_{\alpha, \beta} y$ or there exists a sequence of elements $u_1, u_2, u_3, \cdots, u_n$ in $U$ such that $x R_{\alpha, \beta} u_1, u_1 R_{\alpha, \beta} u_2, u_2 R_{\alpha, \beta} u_3, \cdots, u_n R_{\alpha, \beta} y$. In the last case, we say that $x$ is transitively $(\alpha, \beta)$-similar to $y$ with respect to $R$.

It is also easy to see that for any $(\alpha, \beta) \in J$, $R(\alpha, \beta)$ is an equivalence relation on $U$. We denote $R^*_{\alpha, \beta}$ the set of equivalence classes generated by the equivalence relation $R(\alpha, \beta)$ for each fixed $(\alpha, \beta) \in J$.

**Definition 3.6** Let $U$ be a universal set and $R$ be an intuitionistic fuzzy proximity relation on $U$. The pair $(U, R)$ is an intuitionistic fuzzy approximation space (IF-approximation space). An IF-approximation space $(U, R)$ generates usual approximation space $(U, R(\alpha, \beta))$ of Pawlak for every $(\alpha, \beta) \in J$.

**Definition 3.7** The rough set on $X$ in the generalized approximation space $(U, R(\alpha, \beta))$ is denoted by $(\underline{X}_{\alpha,\beta}, \overline{X}_{\alpha,\beta})$, where

$$\underline{X}_{\alpha,\beta} = \bigcup\{Y | Y \in R^*_{\alpha,\beta} \text{ and } Y \subseteq X\} \quad (5)$$

and $\overline{X}_{\alpha,\beta} = \bigcup\{Y | Y \in R^*_{\alpha,\beta} \text{ and } Y \cap X \neq \phi\} \quad (6)$

**Definition 3.8** Let $X$ be a rough set in the generalized approximation space $(U, R(\alpha, \beta))$. Then we define the $(\alpha, \beta)$-boundary of $X$ with respect to $R$ denoted by $BNR_{\alpha,\beta}(X)$ as $BNR_{\alpha,\beta}(X) = \overline{X}_{\alpha,\beta} - \underline{X}_{\alpha,\beta}$.

**Definition 3.9** Let $X$ be a rough set in the generalized approximation space $(U, R(\alpha, \beta))$. Then $X$ is $(\alpha, \beta)$-*discernible* with respect to $R$ if and only if $\overline{X}_{\alpha,\beta} = \underline{X}_{\alpha,\beta}$ and $X$ is $(\alpha, \beta)$-*rough* with respect to $R$ if and only if $\overline{X}_{\alpha,\beta} \neq \underline{X}_{\alpha,\beta}$.

## 4. Order Information Table

The basic objective of inductive learning and data mining is to learn the knowledge for classification. It is certainly true for rough set theory based approaches. However, in real world problems, we may not be faced with simply classification. One such problem is the ordering of objects. We are interested in mining the association between the overall ordering and the individual orderings induced by different attributes. In many information systems, a set of objects are typically represented by their values on a finite set of attributes. Such information may be conveniently described in a tabular form. Formally, an information table is defined as a quadruple:

$$I = (U, A, \{V_a : a \in A\}, \{f_a : a \in A\}),$$

where $U$ is a finite non-empty set of objects called the universe and $A$ is a non-empty finite set of attributes. For every $a \in A$, $V_a$ is the set of values that attribute $a$ may take and $f_a : U \to V_a$ is an information function [2]. In practical applications, there are various possible interpretations of objects such as cases, states, patients, processes, and observations. Attributes can be interpreted as features, variables, and characteristics. A special case of information systems called information table where the columns are labeled by attributes and rows are by objects. For example: The information table assigns a particular value $a(x)$ from $V_a$ to each attribute $a$ and object $x$ in the universe $U$. With any $P \subseteq A$ there is an associated equivalence relation $IND(P)$ such that

$$IND(P) = \{(x, y) \in U^2 \mid \forall \ a \in P, a(x) = a(y)\}$$

The relation $IND(P)$ is called a P-indiscernibility relation. The partition of $U$ is a family of all equivalence classes of $IND(P)$ and is denoted by $U/IND(P)$ or $U/P$. If $(x, y) \in IND(P)$, then $x$ and $y$ are indiscernible by attributes from $P$. For example, consider the information Table 1, where the attribute $a_1$ represents the availability of research and development facility; $a_2$ represents the adoption of state of art facility; $a_3$ represents the marketing expenses, and $a_4$ represents the profits in million of rupees.

. Table 1. Information table

| Object | R&D facility ($a_1$) | State of art facility ($a_2$) | Marketing expenses ($a_3$) | Profits in million Rs. ($a_4$) |
|---|---|---|---|---|
| $o_1$ | No | Yes | High | 200 |
| $o_2$ | Yes | No | High | 300 |
| $o_3$ | Yes | Yes | Average | 200 |
| $o_4$ | No | Yes | Very high | 250 |
| $o_5$ | Yes | No | High | 300 |
| $o_6$ | No | Yes | Very high | 250 |

An ordered information table (OIT) is defined as $\text{OIT} = \{\text{IT}, \{\prec_a : a \in A\}\}$ where, IT is a standard information table and $\prec_a$ is an order relation on attribute $a$. An ordering of values of a particular attribute a naturally induces an ordering of objects:

$$x \prec_{\{a\}} y \Leftrightarrow f_a(x) \prec_a f_a(y)$$

where, $\prec_{\{a\}}$ denotes an order relation on $U$ induced by the attribute $a$. An object $x$ is ranked ahead of object $y$ if and only if the value of $x$ on the attribute $a$ is ranked ahead of the value of $y$ on the attribute $a$. For example, information Table 1 is considered as ordered information table if

$\prec_{a_1}$: Yes $\prec$ No

$\prec_{a_2}$: Yes $\prec$ No

$\prec_{a_3}$: Very high $\prec$ High $\prec$ Average

$\prec_{a_4}$: 300 $\prec$ 250 $\prec$ 200

For a subset of attributes $P \subseteq A$, we define:

$$x \prec_P y \Leftrightarrow f_a(x) \prec_a f_a(y) \quad \forall a \in P$$
$$\Leftrightarrow \bigwedge_{a \in P} f_a(x) \prec_a f_a(y)$$
$$\Leftrightarrow \bigcap_{a \in P} \prec_{\{a\}}$$

In practical applications, there are various possible interpretations of objects such as cases, states, patients,

processes, and observations. Attributes can be interpreted as features, variables, and characteristics. A special case of information systems called information table where the columns are labeled by attributes and rows are by objects. For example: However in real life applications, it is observed that the attribute values are not exactly identical but almost identical. This is because objects characterized by the almost same information are almost indiscernible in the view of available information. At this point we generalize Pawlak's approach of indiscernibility. Keeping view to this, the almost indiscernibility relation generated in this way is the mathematical basis of rough set on fuzzy approximation space and is discussed by Tripathy and Acharjya [4]. This is further generalized to rough set on intuitionistic fuzzy approximation space by Tripathy [1]. Also, generalized information table may be viewed as information tables with added semantics. For the problem of knowledge mining, we introduce order relations on attribute values [2]. However, it is not appropriate in case of attribute values that are almost indiscernible.

In this paper we use rough sets on intuitionistic fuzzy approximation space to find the attribute values that are $(\alpha, \beta)$-identical before introducing the order relation. This is because exactly ordering is not possible when the attribute values are almost identical. For $\alpha = 1, \beta = 0$, the almost indiscernibility relation, reduces to the indiscernibility relation. Therefore, it generalizes the Pawlak's indiscernibility relation.

## 5. Basics of Formal Concept Analysis

Formal Concept Analysis has been introduced by R. Wille [8] is a method of analyzing data across various domains such as psychology, sociology, anthropology, medicine, biology, linguistics, computer sciences and industrial engineering to name a few. The basic aim of this theory is to construct a concept lattice that provides complete information about structure such as relations between object and attributes; implications, and dependencies. At the same time it allows knowledge acquisition from (or by) an expert by putting very precise questions, which either have to be confirmed or to be refuted by a counterexample.

5.1 Formal Context and Formal Concept

In this section we recall the basic definitions and notations of formal concept analysis developed by R. Wille [8]. A formal context is defined as a set structure $K = (G, M, R)$ consists of two sets $G$ and $M$ while $R$ is a binary relation between $G$ and $M$, i.e. $R \subseteq G \times M$. The elements of $G$ are called the objects and the elements of $M$ are called the attributes of the context. The formal concept of the formal context $(G, M, R)$ is defined with the help of derivation operators. The derivation operators are defined for arbitrary $X \subseteq G$ and $Y \subseteq M$ as follows:

$$X' = \{a \in M : uRa \quad \forall \ u \in X\} \tag{7}$$

$$Y' = \{u \in G : uRa \quad \forall \ a \in Y\} \tag{8}$$

A formal concept of a formal context $K = (G, M, R)$ is defined as a pair $(X, Y)$ with $X \subseteq G$, $Y \subseteq M$, $X = Y'$, and $Y = X'$. The first member $X$, of the pair $(X, Y)$ is called the extent whereas the second member $Y$ is called the intent of the formal concept. Objects in $X$ share all properties $Y$, and only properties $Y$ are possessed by all objects in $X$. A basic result is that the formal concepts of a formal context are always forming the mathematical structure of a lattice with respect to the subconcept-superconcept relation. Therefore, the set of all formal concepts forms a complete lattice called a concept lattice [8]. The subconcept-superconcept relation can be best depicted by a lattice diagram and we can derive concepts, implication sets, and association rules based on the cross table. Now we present the cross table of the information table given in Table 1 in Table 2, where the rows are represented as objects and columns are represented as attributes. The relation between them is represented by a cross. The lattice diagram is presented in figure 1.

Table 2 Cross table of the information system

| Object | R&D | | St. of Art | | Marketing | | | Profits | | |
|--------|-----|-----|-----|-----|-----------|------|---------|-----|-----|-----|
|        | Yes | No  | Yes | No  | Very High | High | Average | 200 | 250 | 300 |
| $o_1$  |     | ×   | ×   |     |           | ×    |         | ×   |     |     |
| $o_2$  | ×   |     |     | ×   |           | ×    |         |     |     | ×   |
| $o_3$  | ×   |     | ×   |     |           |      | ×       | ×   |     |     |
| $o_4$  |     | ×   | ×   |     | ×         |      |         |     | ×   |     |
| $o_5$  | ×   |     |     | ×   |           | ×    |         |     |     | ×   |
| $o_6$  |     | ×   | ×   |     | ×         |      |         |     | ×   |     |

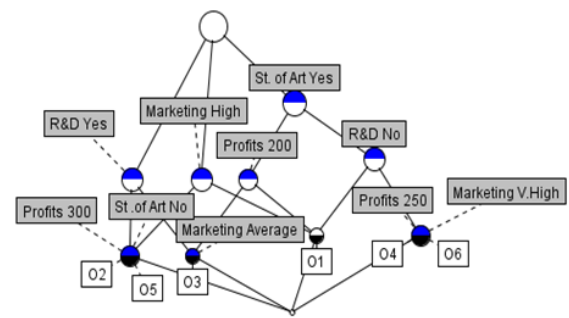

Fig. 1 Lattice diagram of the information system

## 6. Proposed Knowledge Mining Model

In this section, we present proposed knowledge mining model that consists of problem definition, target data, preprocessed data, processed data, data classification, ordered information table, knowledge, and computation of the chief attribute as shown in Figure 2. Problem definition and acquiring of prior knowledge are the fundamental steps of any model when the right problem is identified. The pattern of data elements or the usefulness of the individual data element change dramatically from individual to individual, organization to organization, or task to task because of the acquisition of knowledge and reasoning may involve vagueness and incompleteness.

It's difficult to obtain the useful information that is hidden in the huge database by an individual. Therefore, it is essential to deal with incomplete and vague information in classification, concept formulation, and data analysis present in the database. In this model we use intuitionistic fuzzy proximity relation, ordering rules, and formal concept analysis to identify the chief attribute instead of only fuzzy proximity relation as discussed in [2]. In the preprocess we use intuitionistic fuzzy proximity relation as it provides better knowledge over fuzzy proximity relation as stated in [3]. Finally, formal concept analysis on these knowledge are inclined to explore better knowledge and to find out most important factors affecting the decision making process.

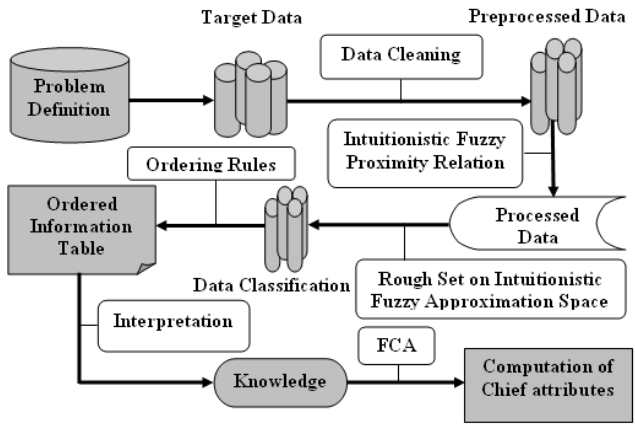

Fig. 2 Proposed knowledge mining model

## 7. Study on Ranking Institutions

For demonstration of our model we take into consideration an information system of a group of institutions of any country where we study the ranking of institution and the parameters influencing the rank. In Table 3 below, we specify the attribute descriptions that influence the ranking of the institutions. The institutes can be judged by the outputs, which are produced. The quality of the output can be judged by the placement performance of the institute and is given highest weight with a score 385 which comes to around 24% of total weight. To produce the quality output the input should be of high quality. The major inputs for an institute are in general intellectual capital and infrastructure facilities. Accordingly the scores for intellectual capital and an infrastructure facility are fixed as 250 and 200 respectively that weight 15% and 12% of total weight. The student placed in the company shall serve the company up to their expectation and it leads to recruiter's satisfaction and is given with a score of 200 which comes around 12%. At the same time students satisfaction and extra curricular activities plays a vital role for prospective students is given with a score 60 and 80 respectively of weight 4% and 6% of the total weight. We have not considered many other factors that do not have impact on ranking the institutions and to make our analysis simple

The membership and non-membership functions have been adjusted such that the sum of their values should lie in [0, 1] and also these functions must be symmetric. The first requirement necessitates a major of 2 in the denominators of the non-membership functions.

Table 3 Attribute descriptions table

| Attribute | Notation | Possible Range | Membership Function | Non Membership function |
|---|---|---|---|---|
| Intellectual capital (IC) | $A_1$ | 1 - 250 | $1 - \frac{|x-y|}{250}$ | $\frac{|x-y|}{2(x+y)}$ |
| Infrastructure facility (IF) | $A_2$ | 1 - 200 | $1 - \frac{|x-y|}{200}$ | $\frac{|x-y|}{2(x+y)}$ |
| Placement performance (PP) | $A_3$ | 1 - 385 | $1 - \frac{|x-y|}{385}$ | $\frac{|x-y|}{2(x+y)}$ |
| Recruiters satisfaction score (RS) | $A_4$ | 1 - 200 | $1 - \frac{|x-y|}{200}$ | $\frac{|x-y|}{2(x+y)}$ |
| Students satisfaction score (SS) | $A_5$ | 1 - 60 | $1 - \frac{|x-y|}{60}$ | $\frac{|x-y|}{2(x+y)}$ |
| Extra curricular activities (ECA) | $A_6$ | 1- 80 | $1 - \frac{|x-y|}{80}$ | $\frac{|x-y|}{2(x+y)}$ |

We consider a small universe of 10 institutions and information related to it are given in the following Table 4.

Table 4 Small universe of information system

| Name of the Institute | IC | IF | PP | RS | SS | ECA |
|---|---|---|---|---|---|---|
| $i_1$ | 229 | 151 | 304 | 169 | 56 | 49 |
| $i_2$ | 227 | 143 | 298 | 169 | 53 | 79 |
| $i_3$ | 226 | 145 | 266 | 167 | 54 | 63 |
| $i_4$ | 191 | 110 | 316 | 163 | 41 | 64 |
| $i_5$ | 179 | 117 | 247 | 160 | 53 | 53 |
| $i_6$ | 148 | 102 | 180 | 147 | 43 | 27 |
| $i_7$ | 131 | 78 | 138 | 145 | 46 | 25 |
| $i_8$ | 124 | 61 | 130 | 142 | 38 | 9 |
| $i_9$ | 88 | 58 | 121 | 143 | 40 | 34 |
| $i_{10}$ | 92 | 48 | 100 | 137 | 32 | 2 |

Intuitionistic fuzzy proximity relation $R_1$ corresponding to attribute 'IC' is given in Table 5.

Table 5  Intuitionistic fuzzy proximity relation for attribute IC

| Name of the Institute | $i_1$ | $i_2$ | $i_3$ | $i_4$ | $i_5$ | $i_6$ | $i_7$ | $i_8$ | $i_9$ | $i_{10}$ |
|---|---|---|---|---|---|---|---|---|---|---|
| $i_1$ | 1, 0 | 0.992, 0.002 | 0.989, 0.003 | 0.848, 0.045 | 0.802, 0.061 | 0.676, 0.108 | 0.61, 0.135 | 0.581, 0.149 | 0.437, 0.222 | 0.451, 0.214 |
| $i_2$ | 0.992, 0.002 | 1, 0 | 0.997, 0.001 | 0.856, 0.043 | 0.81, 0.058 | 0.684, 0.106 | 0.618, 0.133 | 0.589, 0.147 | 0.445, 0.22 | 0.46, 0.212 |
| $i_3$ | 0.989, 0.003 | 0.997, 0.001 | 1, 0 | 0.859, 0.042 | 0.813, 0.058 | 0.686, 0.105 | 0.621, 0.132 | 0.591, 0.146 | 0.448, 0.219 | 0.462, 0.211 |
| $i_4$ | 0.848, 0.045 | 0.856, 0.043 | 0.859, 0.042 | 1, 0 | 0.954, 0.016 | 0.827, 0.064 | 0.762, 0.092 | 0.732, 0.106 | 0.589, 0.184 | 0.603, 0.175 |
| $i_5$ | 0.802, 0.061 | 0.81, 0.058 | 0.813, 0.058 | 0.954, 0.016 | 1, 0 | 0.873, 0.048 | 0.808, 0.077 | 0.778, 0.091 | 0.635, 0.17 | 0.649, 0.162 |
| $i_6$ | 0.676, 0.108 | 0.684, 0.106 | 0.686, 0.105 | 0.827, 0.064 | 0.873, 0.048 | 1, 0 | 0.935, 0.029 | 0.905, 0.044 | 0.762, 0.126 | 0.776, 0.117 |
| $i_7$ | 0.61, 0.135 | 0.618, 0.133 | 0.621, 0.132 | 0.762, 0.092 | 0.808, 0.077 | 0.935, 0.029 | 1, 0 | 0.97, 0.014 | 0.827, 0.098 | 0.841, 0.089 |
| $i_8$ | 0.581, 0.149 | 0.589, 0.147 | 0.591, 0.146 | 0.732, 0.106 | 0.778, 0.091 | 0.905, 0.044 | 0.97, 0.014 | 1, 0 | 0.857, 0.084 | 0.871, 0.075 |
| $i_9$ | 0.437, 0.222 | 0.445, 0.22 | 0.448, 0.219 | 0.589, 0.184 | 0.635, 0.17 | 0.762, 0.126 | 0.827, 0.098 | 0.857, 0.084 | 1, 0 | 0.986, 0.01 |
| $i_{10}$ | 0.451, 0.214 | 0.46, 0.212 | 0.462, 0.211 | 0.603, 0.175 | 0.649, 0.162 | 0.776, 0.117 | 0.841, 0.089 | 0.871, 0.075 | 0.986, 0.01 | 1, 0 |

Intuitionistic fuzzy proximity relation $R_2$ corresponding to attribute 'IF' is given in Table 6.

Table 6  Intuitionistic fuzzy proximity relation for attribute IF

| Name of the Institute | $i_1$ | $i_2$ | $i_3$ | $i_4$ | $i_5$ | $i_6$ | $i_7$ | $i_8$ | $i_9$ | $i_{10}$ |
|---|---|---|---|---|---|---|---|---|---|---|
| $i_1$ | 1,0 | 0.956, 0.015 | 0.966, 0.011 | 0.792, 0.08 | 0.827, 0.065 | 0.751, 0.098 | 0.632, 0.161 | 0.549, 0.212 | 0.532, 0.224 | 0.485, 0.258 |
| $i_2$ | 0.956, 0.015 | 1,0 | 0.99, 0.004 | 0.836, 0.065 | 0.871, 0.05 | 0.795, 0.084 | 0.676, 0.147 | 0.593, 0.2 | 0.576, 0.212 | 0.529, 0.246 |
| $i_3$ | 0.966, 0.011 | 0.99, 0.004 | 1,0 | 0.825, 0.069 | 0.86, 0.054 | 0.785, 0.087 | 0.666, 0.15 | 0.583, 0.203 | 0.565, 0.215 | 0.519, 0.249 |
| $i_4$ | 0.792, 0.08 | 0.836, 0.065 | 0.825, 0.069 | 1,0 | 0.965, 0.015 | 0.96, 0.019 | 0.84, 0.085 | 0.757, 0.142 | 0.74, 0.155 | 0.693, 0.194 |
| $i_5$ | 0.827, 0.065 | 0.871, 0.05 | 0.86, 0.054 | 0.965, 0.015 | 1,0 | 0.925, 0.034 | 0.806, 0.1 | 0.723, 0.156 | 0.705, 0.169 | 0.659, 0.207 |
| $i_6$ | 0.751, 0.098 | 0.795, 0.084 | 0.785, 0.087 | 0.96, 0.019 | 0.925, 0.034 | 1,0 | 0.881, 0.066 | 0.798, 0.124 | 0.78, 0.138 | 0.734, 0.177 |
| $i_7$ | 0.632, 0.161 | 0.676, 0.147 | 0.666, 0.15 | 0.84, 0.085 | 0.806, 0.1 | 0.881, 0.066 | 1,0 | 0.917, 0.06 | 0.9, 0.074 | 0.853, 0.116 |
| $i_8$ | 0.549, 0.212 | 0.593, 0.2 | 0.583, 0.203 | 0.757, 0.142 | 0.723, 0.156 | 0.798, 0.124 | 0.917, 0.06 | 1,0 | 0.983, 0.015 | 0.936, 0.058 |
| $i_9$ | 0.532, 0.224 | 0.576, 0.212 | 0.565, 0.215 | 0.74, 0.155 | 0.705, 0.169 | 0.78, 0.138 | 0.9, 0.074 | 0.983, 0.015 | 1,0 | 0.954, 0.044 |
| $i_{10}$ | 0.485, 0.258 | 0.529, 0.246 | 0.519, 0.249 | 0.693, 0.194 | 0.659, 0.207 | 0.734, 0.177 | 0.853, 0.116 | 0.936, 0.058 | 0.954, 0.044 | 1,0 |

Intuitionistic fuzzy proximity relation $R_3$ corresponding to attribute 'PP' is given in Table 7.

Table 7  Intuitionistic fuzzy proximity relation for attribute PP

| Name of the Institute | $i_1$ | $i_2$ | $i_3$ | $i_4$ | $i_5$ | $i_6$ | $i_7$ | $i_8$ | $i_9$ | $i_{10}$ |
|---|---|---|---|---|---|---|---|---|---|---|
| $i_1$ | 1, 0 | 0.987, 0.004 | 0.904, 0.033 | 0.967, 0.01 | 0.853, 0.052 | 0.678, 0.128 | 0.57, 0.187 | 0.55, 0.2 | 0.525, 0.215 | 0.472, 0.252 |
| $i_2$ | 0.987, 0.004 | 1, 0 | 0.917, 0.028 | 0.954, 0.015 | 0.866, 0.047 | 0.691, 0.124 | 0.583, 0.184 | 0.563, 0.196 | 0.539, 0.212 | 0.485, 0.249 |
| $i_3$ | 0.904, 0.033 | 0.917, 0.028 | 1, 0 | 0.87, 0.043 | 0.949, 0.019 | 0.774, 0.097 | 0.666, 0.159 | 0.646, 0.172 | 0.622, 0.188 | 0.568, 0.227 |
| $i_4$ | 0.967, 0.01 | 0.954, 0.015 | 0.87, 0.043 | 1, 0 | 0.819, 0.062 | 0.645, 0.138 | 0.537, 0.196 | 0.517, 0.208 | 0.492, 0.224 | 0.439, 0.259 |
| $i_5$ | 0.853, 0.052 | 0.866, 0.047 | 0.949, 0.019 | 0.819, 0.062 | 1, 0 | 0.825, 0.079 | 0.717, 0.141 | 0.697, 0.154 | 0.673, 0.171 | 0.619, 0.211 |
| $i_6$ | 0.678, 0.128 | 0.691, 0.124 | 0.774, 0.097 | 0.645, 0.138 | 0.825, 0.079 | 1, 0 | 0.892, 0.065 | 0.872, 0.079 | 0.847, 0.098 | 0.794, 0.142 |
| $i_7$ | 0.57, 0.187 | 0.583, 0.184 | 0.666, 0.159 | 0.537, 0.196 | 0.717, 0.141 | 0.892, 0.065 | 1, 0 | 0.98, 0.014 | 0.955, 0.033 | 0.902, 0.079 |
| $i_8$ | 0.55, 0.2 | 0.563, 0.196 | 0.646, 0.172 | 0.517, 0.208 | 0.697, 0.154 | 0.872, 0.079 | 0.98, 0.014 | 1, 0 | 0.975, 0.019 | 0.922, 0.065 |
| $i_9$ | 0.525, 0.215 | 0.539, 0.212 | 0.622, 0.188 | 0.492, 0.224 | 0.673, 0.171 | 0.847, 0.098 | 0.955, 0.033 | 0.975, 0.019 | 1, 0 | 0.947, 0.046 |
| $i_{10}$ | 0.472, 0.252 | 0.485, 0.249 | 0.568, 0.227 | 0.439, 0.259 | 0.619, 0.211 | 0.794, 0.142 | 0.902, 0.079 | 0.922, 0.065 | 0.947, 0.046 | 1, 0 |

Intuitionistic fuzzy proximity relation $R_4$ corresponding to attribute 'RS' is given in Table 8.

Table 8. Intuitionistic fuzzy proximity relation for attribute RS

| Name of the Institute | $i_1$ | $i_2$ | $i_3$ | $i_4$ | $i_5$ | $i_6$ | $i_7$ | $i_8$ | $i_9$ | $i_{10}$ |
|---|---|---|---|---|---|---|---|---|---|---|
| $i_1$ | 1, 0 | 1, 0 | 0.99, 0.003 | 0.97, 0.009 | 0.955, 0.014 | 0.89, 0.035 | 0.88, 0.038 | 0.865, 0.043 | 0.87, 0.042 | 0.84, 0.052 |
| $i_2$ | 1, 0 | 1, 0 | 0.99, 0.003 | 0.97, 0.009 | 0.955, 0.014 | 0.89, 0.035 | 0.88, 0.038 | 0.865, 0.043 | 0.87, 0.042 | 0.84, 0.052 |
| $i_3$ | 0.99, 0.003 | 0.99, 0.003 | 1, 0 | 0.98, 0.006 | 0.965, 0.011 | 0.9, 0.032 | 0.89, 0.035 | 0.875, 0.04 | 0.88, 0.039 | 0.85, 0.049 |
| $i_4$ | 0.97, 0.009 | 0.97, 0.009 | 0.98, 0.006 | 1, 0 | 0.985, 0.005 | 0.92, 0.026 | 0.91, 0.029 | 0.895, 0.034 | 0.9, 0.033 | 0.87, 0.043 |
| $i_5$ | 0.955, 0.014 | 0.955, 0.014 | 0.965, 0.011 | 0.985, 0.005 | 1, 0 | 0.935, 0.021 | 0.925, 0.025 | 0.91, 0.03 | 0.915, 0.028 | 0.885, 0.039 |
| $i_6$ | 0.89, 0.035 | 0.89, 0.035 | 0.9, 0.032 | 0.92, 0.026 | 0.935, 0.021 | 1, 0 | 0.99, 0.003 | 0.975, 0.009 | 0.98, 0.007 | 0.95, 0.018 |
| $i_7$ | 0.88, 0.038 | 0.88, 0.038 | 0.89, 0.035 | 0.91, 0.029 | 0.925, 0.025 | 0.99, 0.003 | 1, 0 | 0.985, 0.005 | 0.99, 0.003 | 0.96, 0.014 |
| $i_8$ | 0.865, 0.043 | 0.865, 0.043 | 0.875, 0.04 | 0.895, 0.034 | 0.91, 0.03 | 0.975, 0.009 | 0.985, 0.005 | 1, 0 | 0.995, 0.002 | 0.975, 0.009 |
| $i_9$ | 0.87, 0.042 | 0.87, 0.042 | 0.88, 0.039 | 0.9, 0.033 | 0.915, 0.028 | 0.98, 0.007 | 0.99, 0.003 | 0.995, 0.002 | 1, 0 | 0.97, 0.011 |
| $i_{10}$ | 0.84, 0.052 | 0.84, 0.052 | 0.85, 0.049 | 0.87, 0.043 | 0.885, 0.039 | 0.95, 0.018 | 0.96, 0.014 | 0.975, 0.009 | 0.97, 0.011 | 1, 0 |

Intuitionistic fuzzy proximity relation $R_5$ corresponding to the attribute 'SS' is given in Table 9.

Table 9  Intuitionistic fuzzy proximity relation for attribute SS

| Name of the Institute | $i_1$ | $i_2$ | $i_3$ | $i_4$ | $i_5$ | $i_6$ | $i_7$ | $i_8$ | $i_9$ | $i_{10}$ |
|---|---|---|---|---|---|---|---|---|---|---|
| $i_1$ | 1, 0 | 0.95, 0.014 | 0.967, 0.009 | 0.75, 0.077 | 0.95, 0.014 | 0.783, 0.066 | 0.833, 0.049 | 0.7, 0.096 | 0.733, 0.083 | 0.6, 0.136 |
| $i_2$ | 0.95, 0.014 | 1, 0 | 0.983, 0.005 | 0.8, 0.064 | 1, 0 | 0.833, 0.052 | 0.883, 0.035 | 0.75, 0.082 | 0.783, 0.07 | 0.65, 0.124 |
| $i_3$ | 0.967, 0.009 | 0.983, 0.005 | 1, 0 | 0.783, 0.068 | 0.983, 0.005 | 0.817, 0.057 | 0.867, 0.04 | 0.733, 0.087 | 0.767, 0.074 | 0.633, 0.128 |
| $i_4$ | 0.75, 0.077 | 0.8, 0.064 | 0.783, 0.068 | 1, 0 | 0.8, 0.064 | 0.967, 0.012 | 0.917, 0.029 | 0.95, 0.019 | 0.983, 0.006 | 0.85, 0.062 |
| $i_5$ | 0.95, 0.014 | 1, 0 | 0.983, 0.005 | 0.8, 0.064 | 1, 0 | 0.833, 0.052 | 0.883, 0.035 | 0.75, 0.082 | 0.783, 0.07 | 0.65, 0.124 |
| $i_6$ | 0.783, 0.066 | 0.833, 0.052 | 0.817, 0.057 | 0.967, 0.012 | 0.833, 0.052 | 1, 0 | 0.95, 0.017 | 0.917, 0.031 | 0.95, 0.018 | 0.817, 0.073 |
| $i_7$ | 0.833, 0.049 | 0.883, 0.035 | 0.867, 0.04 | 0.917, 0.029 | 0.883, 0.035 | 0.95, 0.017 | 1, 0 | 0.867, 0.048 | 0.9, 0.035 | 0.767, 0.09 |
| $i_8$ | 0.7, 0.096 | 0.75, 0.082 | 0.733, 0.087 | 0.95, 0.019 | 0.75, 0.082 | 0.917, 0.031 | 0.867, 0.048 | 1, 0 | 0.967, 0.013 | 0.9, 0.043 |
| $i_9$ | 0.733, 0.083 | 0.783, 0.07 | 0.767, 0.074 | 0.983, 0.006 | 0.783, 0.07 | 0.95, 0.018 | 0.9, 0.035 | 0.967, 0.013 | 1, 0 | 0.867, 0.056 |
| $i_{10}$ | 0.6, 0.136 | 0.65, 0.124 | 0.633, 0.128 | 0.85, 0.062 | 0.65, 0.124 | 0.817, 0.073 | 0.767, 0.09 | 0.9, 0.043 | 0.867, 0.056 | 1, 0 |

Intuitionistic fuzzy proximity relation $R_6$ corresponding to attribute 'ECA' is given in Table 10.

Table 10  Intuitionistic fuzzy proximity relation for attribute ECA

| Name of the Institute | $i_1$ | $i_2$ | $i_3$ | $i_4$ | $i_5$ | $i_6$ | $i_7$ | $i_8$ | $i_9$ | $i_{10}$ |
|---|---|---|---|---|---|---|---|---|---|---|
| $i_1$ | 1, 0 | 0.631, 0.115 | 0.831, 0.06 | 0.821, 0.063 | 0.953, 0.018 | 0.726, 0.143 | 0.692, 0.167 | 0.499, 0.342 | 0.812, 0.09 | 0.407, 0.462 |
| $i_2$ | 0.631, 0.115 | 1, 0 | 0.8, 0.056 | 0.81, 0.053 | 0.679, 0.097 | 0.357, 0.242 | 0.323, 0.262 | 0.13, 0.395 | 0.443, 0.197 | 0.039, 0.476 |
| $i_3$ | 0.831, 0.06 | 0.8, 0.056 | 1, 0 | 0.99, 0.003 | 0.879, 0.042 | 0.557, 0.196 | 0.523, 0.218 | 0.33, 0.372 | 0.643, 0.147 | 0.239, 0.47 |
| $i_4$ | 0.821, 0.063 | 0.81, 0.053 | 0.99, 0.003 | 1, 0 | 0.869, 0.045 | 0.547, 0.199 | 0.513, 0.221 | 0.32, 0.373 | 0.633, 0.15 | 0.229, 0.47 |
| $i_5$ | 0.953, 0.018 | 0.679, 0.097 | 0.879, 0.042 | 0.869, 0.045 | 1, 0 | 0.679, 0.16 | 0.644, 0.183 | 0.452, 0.352 | 0.764, 0.108 | 0.36, 0.465 |
| $i_6$ | 0.726, 0.143 | 0.357, 0.242 | 0.557, 0.196 | 0.547, 0.199 | 0.679, 0.16 | 1, 0 | 0.966, 0.026 | 0.773, 0.248 | 0.914, 0.056 | 0.681, 0.434 |
| $i_7$ | 0.692, 0.167 | 0.323, 0.262 | 0.523, 0.218 | 0.513, 0.221 | 0.644, 0.183 | 0.966, 0.026 | 1, 0 | 0.807, 0.227 | 0.88, 0.081 | 0.716, 0.427 |
| $i_8$ | 0.499, 0.342 | 0.13, 0.395 | 0.33, 0.372 | 0.32, 0.373 | 0.452, 0.352 | 0.773, 0.248 | 0.807, 0.227 | 1, 0 | 0.687, 0.287 | 0.909, 0.327 |
| $i_9$ | 0.812, 0.09 | 0.443, 0.197 | 0.643, 0.147 | 0.633, 0.15 | 0.764, 0.108 | 0.914, 0.056 | 0.88, 0.081 | 0.687, 0.287 | 1, 0 | 0.596, 0.446 |
| $i_{10}$ | 0.407, 0.462 | 0.039, 0.476 | 0.239, 0.47 | 0.229, 0.47 | 0.36, 0.465 | 0.681, 0.434 | 0.716, 0.427 | 0.909, 0.327 | 0.596, 0.446 | 1, 0 |

On considering the degree of dependency values $\alpha \geq 0.92$, $\beta < 0.08$ for membership and non-membership functions respectively we have obtained the following equivalence classes.

$U/R_1^{\alpha,\beta} = \{\{i_1,i_2,i_3\},\{i_4,i_5\},\{i_6,i_7,i_8\},\{i_9,i_{10}\}\}$

$U/R_2^{\alpha,\beta} = \{\{i_1,i_2,i_3\},\{i_4,i_5,i_6\},\{i_7\},\{i_8,i_9,i_{10}\}\}$

$U/R_3^{\alpha,\beta} = \{\{i_1,i_2,i_4\},\{i_3,i_5\},\{i_6\},\{i_7,i_8,i_9,i_{10}\}\}$

$U/R_4^{\alpha,\beta} = \{\{i_1,i_2,i_3,i_4,i_5,i_6,i_7,i_8,i_9,i_{10}\}\}$

$U/R_5^{\alpha,\beta} = \{\{i_1,i_2,i_3,i_5\},\{i_4,i_6,i_7,i_8,i_9\},\{i_{10}\}\}$

$U/R_6^{\alpha,\beta} = \{\{i_1,i_5\},\{i_2\},\{i_3,i_4\},\{i_6,i_7\},\{i_8\},\{i_9,i_{10}\}\}$

From the above analysis, it is clear that the attribute ECA classify the universe into six categories. Let it be low, average, good, very good, excellent, and outstanding and hence can be ordered. Similarly, the attributes IC, IF, and PP classify the universe into four categories. Let it be low, moderate, high and very high and hence can be ordered. The attribute SS classify the universe into three categories namely good, very good, and excellent. Since the equivalence class $U/R_4^{\alpha,\beta}$ contains only one group, the universe is $(\alpha,\beta)$-indiscernible according to the attribute RS and hence do not require any ordering while extracting knowledge from the information system. Therefore, the ordered information table of the small universe Table 4 is given in Table 11.

Table 11 Ordered information table of the small universe

| Institutions | IC | IF | PP | SS | ECA |
|---|---|---|---|---|---|
| $i_1$ | Very high | Very high | Very high | Excellent | Outstanding |
| $i_2$ | Very high | Very high | Very high | Excellent | Excellent |
| $i_3$ | Very high | Very high | High | Excellent | Very good |
| $i_4$ | High | High | Very high | Very good | Very good |
| $i_5$ | High | High | High | Excellent | Outstanding |
| $i_6$ | Moderate | High | Moderate | Very good | Good |
| $i_7$ | Moderate | Moderate | Low | Very good | Good |
| $i_8$ | Moderate | Low | Low | Very good | Average |
| $i_9$ | Low | Low | Low | Very good | Poor |
| $i_{10}$ | Low | Low | Low | Good | Poor |

$\prec_{IC}$: Very high $\prec$ High $\prec$ Moderate $\prec$ Low

$\prec_{IF}$: Very high $\prec$ High $\prec$ Moderate $\prec$ Low

$\prec_{PP}$: Very high $\prec$ High $\prec$ Moderate $\prec$ Low

$\prec_{SS}$: Excellent $\prec$ Very good $\prec$ Good

$\prec_{ECA}$: Outstanding $\prec$ Excellent $\prec$ Very good $\prec$ Good $\prec$ Average $\prec$ Poor

Now, in order to rank the institutions we assign weights to the attribute values. In order to compute the rank of the institutions $i_k; k = 1,2,\cdots,10$ we add the weights of the attribute values and rank them according to the total sum obtained from highest to lowest. However, it is identified that in some cases the total sum remains same for certain institutions. It indicates that these institutions cannot be distinguished from one another according to the available attributes and attribute values. In such cases, using further analysis techniques actual ranking of the institutes can also be found out. On considering the weights of outstanding, excellent, (very high, very good), (high, good), (moderate, average) and (low, poor) as 6, 5, 4, 3, 2 and 1 respectively the ordered information table for ranking the institutions is given in table 12.

Table 12 Ordered Information table for ranking institution

| Institutions | IC | IF | PP | SS | ECA | Total sum | Rank |
|---|---|---|---|---|---|---|---|
| $i_1$ | Very high (4) | Very high (4) | Very high (4) | Excellent (5) | Outstanding (6) | 23 | 1 |
| $i_2$ | Very high (4) | Very high (4) | Very high (4) | Excellent (5) | Excellent (5) | 22 | 2 |
| $i_3$ | Very High (4) | Very high (4) | High (3) | Excellent (5) | Very good (4) | 20 | 3 |
| $i_4$ | High (3) | High (3) | Very high (4) | Very good (4) | Very good (4) | 18 | 4 |
| $i_5$ | High (3) | High (3) | High (3) | Excellent (5) | Outstanding (6) | 20 | 3 |
| $i_6$ | Moderate (2) | High (3) | Moderate (2) | Very good (4) | Good (3) | 14 | 5 |
| $i_7$ | Moderate (2) | Moderate (2) | Low (1) | Very good (4) | Good (3) | 12 | 6 |
| $i_8$ | Moderate (2) | Low (1) | Low (1) | Very good (4) | Average (2) | 10 | 7 |
| $i_9$ | Low (1) | Low (1) | Low (1) | Very good (4) | Poor (1) | 8 | 8 |
| $i_{10}$ | Low (1) | Low (1) | Low (1) | Good (3) | Poor (1) | 7 | 9 |

From the computation given in Table 12 shows that the institution $i_1$ belongs to the first rank whereas $i_3$ and $i_5$ belongs to third rank. Similarly, the ranks of the other institutions can also be obtained from the Table 12.

In the second part of post process initially, we group the institutions based on their rank. Since the difference of the total sum between the first three ranks is much less, we combine these four institutions into a single cluster 1 for our further analysis. Similarly, we combine the institutions having rank 4, 5, and 6 as another cluster 2. The rest of the institutions are combined and is the final cluster 3 for our analysis. Formal concept analysis can do the data classification. However, data was already classified in this study. The purpose of this research is to use formal concept analysis to diagnose the relationship among attributes belonging to the different clusters. Now we present the context table in Table 13 and in figure 3, the lattice diagram for order information Table 12.

Table 13 Context table for order information table

| Institutions | $A_1$ | | | | $A_2$ | | | | $A_3$ | | | | $A_5$ | | | $A_6$ | | | | | |
|---|---|---|---|---|---|---|---|---|---|---|---|---|---|---|---|---|---|---|---|---|---|
| | $A_{11}$ | $A_{12}$ | $A_{13}$ | $A_{14}$ | $A_{21}$ | $A_{22}$ | $A_{23}$ | $A_{24}$ | $A_{31}$ | $A_{32}$ | $A_{33}$ | $A_{34}$ | $A_{51}$ | $A_{52}$ | $A_{53}$ | $A_{61}$ | $A_{62}$ | $A_{63}$ | $A_{64}$ | $A_{65}$ | $A_{66}$ |
| $i_1$ | x | | | | x | | | | x | | | | x | | | x | | | | | |
| $i_2$ | x | | | | x | | | | x | | | | x | | | | x | | | | |
| $i_3$ | x | | | | x | | | | | x | | | x | | | | | x | | | |
| $i_4$ | | x | | | | x | | | x | | | | | x | | | | x | | | |
| $i_5$ | | x | | | | x | | | | x | | | x | | | x | | | | | |
| $i_6$ | | | x | | | x | | | | | x | | | x | | | | | x | | |
| $i_7$ | | | x | | | | x | | | | | x | | x | | | | | x | | |
| $i_8$ | | | x | | | | | x | | | | x | | x | | | | | | x | |
| $i_9$ | | | | x | | | | x | | | | x | | x | | | | | | | x |
| $i_{10}$ | | | | x | | | | x | | | | x | | | x | | | | | | x |

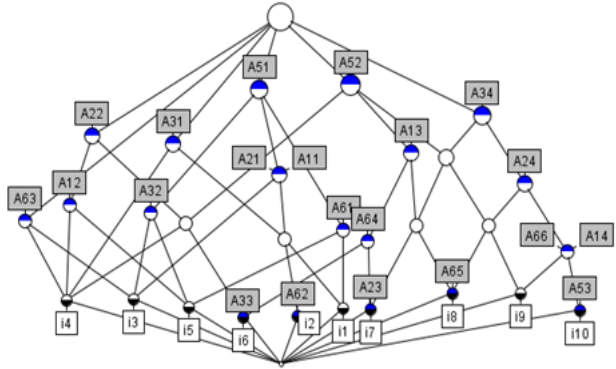

Fig. 3  Lattice diagram for order information table

We generate the implication set for each cluster and are presented in Table 14. Further we find the chief attribute influencing each cluster of institutions by considering the highest frequency obtained in the subconcept superconcept table. For reference, we have presented implication relation Tables 15, 16, and 17 for cluster 1, 2, and 3 respectively.

Table 14  Implication set for all the clusters 1, 2, and 3

| Implication set for ranks 1-3 | |
|---|---|
| 1 < 4 > { } ==> A51; | 6 < 1 > A12 A51 ==> A22 A32 A61; |
| 2 < 3 > A21 A51 ==> A11; | 7 < 1 > A32 A51 A61 ==> A12 A22; |
| 3 < 3 > A11 A51 ==> A21; | 8 < 1 > A11 A21 A51 A61 ==> A31; |
| 4 < 2 > A31 A51 ==> A11 A21; | 9 < 1 > A51 A62 ==> A11 A21 A31; |
| 5 < 1 > A22 A51 ==> A12 A32 A61; | 10 < 1 > A51 A63 ==> A11 A21 A32; |
| | 11 < 1 > A11 A21 A32 A51 ==> A63; |
| **Implication set for ranks 4-6** | |
| 1 < 3 > { } ==> A52; | 6 < 1 > A52 A63 ==> A12 A22 A31; |
| 2 < 2 > A52 A64 ==> A13; | 7 < 1 > A34 A52 ==> A13 A23 A64; |
| 3 < 2 > A13 A52 ==> A64; | 8 < 1 > A33 A52 ==> A13 A22 A64; |
| 4 < 1 > A31 A52 ==> A12 A22 A63; | 9 < 1 > A23 A52 ==> A13 A34 A64; |
| 5 < 1 > A12 A52 ==> A22 A31 A63; | 10 < 1 > A13 A22 A52 A64 ==> A33; |
| **Implication set for ranks 7-9** | |
| 1 < 3 > { } ==> A24 A34; | 4 < 1 > A13 A24 A34 ==> A52 A65; |
| 2 < 2 > A24 A34 A66 ==> A14; | 5 < 1 > A24 A34 A65 ==> A13 A52; |
| 3 < 2 > A14 A24 A34 ==> A66; | 6 < 1 > A24 A34 A53 ==> A14 A66; |

Table 15  Implication relation table for cluster 1

| Imply Superconcept | $A_{11}$ | $A_{12}$ | $A_{21}$ | $A_{22}$ | $A_{31}$ | $A_{32}$ | $A_{51}$ | $A_{61}$ | $A_{63}$ |
|---|---|---|---|---|---|---|---|---|---|
| Subconcept | $A_{21},A_{51}$*3 $A_{31},A_{51}$*2 $A_{51},A_{62}$ $A_{51},A_{63}$ | $A_{22}, A_{51} A_{32}, A_{51} A_{51}, A_{63}$ | $A_{11},A_{51}$*3 $A_{31},A_{51}$*2 $A_{51},A_{62}$ $A_{51},A_{63}$ | $A_{12} A_{51} A_{32} A_{51} A_{61}$ | $A_{12},A_{22}, A_{52} A_{11},A_{21}, A_{51},A_{61} A_{51},A_{62}$ | $A_{22}, A_{51} A_{51}, A_{63}$ | { }*4 | $A_{22},A_{51} A_{12},A_{51}$ | $A_{11},A_{21} A_{32},A_{51}$ |
| Frequency | 14 | 5 | 14 | 5 | 9 | 4 | 0 | 4 | 4 |

Table 16  Implication relation table for cluster 2

| Imply Superconcept | $A_{12}$ | $A_{13}$ | $A_{22}$ | $A_{23}$ | $A_{31}$ | $A_{33}$ | $A_{34}$ | $A_{52}$ | $A_{63}$ | $A_{64}$ |
|---|---|---|---|---|---|---|---|---|---|---|
| Subconcept | $A_{31}, A_{52}$ | $A_{52},A_{64}$*2 | $A_{12}, A_{52}$ | $A_{34}, A_{52}$ | $A_{12}, A_{52}$ | $A_{13}, A_{22}, A_{52}, A_{64}$ | $A_{23}, A_{52}$ | { }*3 | $A_{12}, A_{52}$ | $A_{13},A_{52}$*2 |
| | $A_{52}, A_{63}$ | $A_{23},A_{52} A_{33},A_{52}$ | $A_{33}, A_{52} A_{31}, A_{52}$ | | $A_{52}, A_{63}$ | | | $A_{31} A_{52}$ | | $A_{23},A_{52} A_{33},A_{52}$ |
| | | | $A_{34},A_{52}$ | $A_{52}, A_{63}$ | | | | | | $A_{34},A_{52}$ |
| Frequency | 4 | 10 | 8 | 2 | 4 | 4 | 2 | 0 | 4 | 10 |

Table 17  Implication relation table for cluster 3

| Imply Superconcept | $A_{13}$ | $A_{14}$ | $A_{24}$ | $A_{34}$ | $A_{52}$ | $A_{65}$ | $A_{66}$ |
|---|---|---|---|---|---|---|---|
| Subconcept | $A_{24}, A_{34}, A_{65}$ | $A_{24},A_{34},A_{64}$*2 | { }*3 | { }*3 | $A_{13}, A_{24}, A_{34}$ | $A_{13}, A_{24}, A_{34}$ | $A_{14},A_{24},A_{34}$*2 |
| | | $A_{24},A_{34},A_{53}$ | | | | | $A_{24},A_{34},A_{53}$ |
| Frequency | 3 | 9 | 0 | 0 | 3 | 3 | 9 |

Those highest frequency superconcepts expressed the most important information. In particular, the chief attributes that influence cluster 1 are $A_{11}$ and $A_{21}$ as the frequency is high as obtained in the implication relation table 15. The next influencing factor in the same cluster is $A_{31}$. It indicates that the institutions to be in cluster 1 must be having IC, IF, and PP as very high. Similarly, the influencing factors for cluster 2 obtained from implication table 16 are $A_{13}$ (IC moderate), $A_{64}$ (ECA good), and $A_{22}$ (IF high). Finally, the chief factors that influence cluster 3 obtained from implication Table 17 are $A_{14}$ (IC low), and $A_{66}$ (ECA poor).

## 8. Conclusions

Rough sets on intuitionistic fuzzy approximation spaces introduced in [1], which extends the earlier notion of basic rough sets on fuzzy approximation spaces. Ordering of objects is a fundamental issue in decision making and plays a vital role in the design of intelligent information systems. The main objective of this work is to expand the domain application of rough set on intuitionistic fuzzy approximation space with ordering of objects and to find the chief factors that influence the ranking of the institutions. Our knowledge mining model depicts a layout for performing the ordering of objects using rough sets on intuitionistic fuzzy approximation spaces and the computation of the chief attributes that influence the rank with formal concept analysis. We have taken a real life example of ranking 10 institutes according to different attributes. We have shown how analysis can be done by taking rough set on intuitionistic fuzzy approximation space with ordering rules and formal concept analysis as a model for mining knowledge.

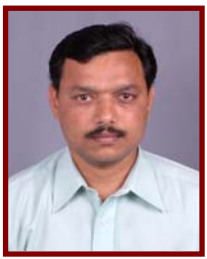
**D. P. Acharjya** received the degree M. Sc. From NIT, Rourkela, India in 1989; M. Phil. from Berhampur University, India in 1994; and M. Tech. degree in computer science from Utkal University, India in 2002. He has been awarded with Gold Medal in M. Sc. He is an associate Professor in School of Computing Sciences and Engineering at VIT University, Vellore, Tamilnadu, India. He has authored many international and national journal papers to his credit. He has also published three books; Fundamental Approach to Discrete Mathematics, Computer Based on Mathematics, and Theory of Computation; to his credit. His research interest includes rough sets, knowledge representation, granular computing, mobile ad-hoc network, and business intelligence. Mr. Acharjya is associated with many professional bodies CSI, ISTE, IMS, AMTI, ISIAM, OITS, IACSIT, IEEE, IAENG, and CSTA. He is the founder secretary of OITS, Rourkela chapter.

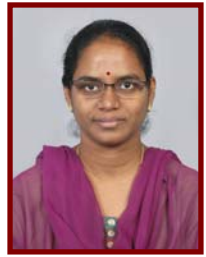
**L. Ezhilarasi** received the M.C.A from Bharathidasan University, Trichy, India in 2000. She is a M. Tech. (CSE) final year student of VIT University, Vellore, India. She has keen interest in teaching and research. Her research interest includes rough sets, fuzzy sets, granular computing, formal concepts, computer graphics and knowledge mining. She is associated with the professional society IRSS.